\documentclass[conference]{IEEEtran}
\IEEEoverridecommandlockouts

\usepackage{cite}
\usepackage{amsmath,amssymb,amsfonts}
\def\BibTeX{{\rm B\kern-.05em{\sc i\kern-.025em b}\kern-.08em
    T\kern-.1667em\lower.7ex\hbox{E}\kern-.125emX}}

\usepackage[utf8]{inputenc} 
\usepackage[T1]{fontenc}    
\usepackage{hyperref}       
\usepackage{url}            
\usepackage{booktabs}       
\usepackage{amsfonts}       
\usepackage{nicefrac}       
\usepackage{microtype}      
\usepackage{xcolor}         
\usepackage{mathtools}
\usepackage{algorithm}
\usepackage{algpseudocode}
\usepackage{amsthm}

\newtheorem{theorem}{Theorem}[section]

\newtheorem{proposition}[theorem]{Proposition}

\newcounter{gaocomm} 
\definecolor{blue-violet}{rgb}{0.00,0.75,0.90} 
\definecolor{bostonuniversityred}{rgb}{1.0, 0.0, 0.0}

\def\X{\mathcal{X}}

\def\RR{\mathbb{R}}

\def\Z{\mathcal{Z}}

\def\bzero{{\mathbf 0}}
\def\bone{{\mathbf 1}}

\def \bp {{\mathbf{p}}}
\def \bq {{\mathbf{q}}}

\def\bx{{\mathbf x}}
\def\by{{\mathbf y}}
\def\bz{{\mathbf z}}

\def\bA{\mathbf A}

\def\bC{\mathbf C}
\def\bD{\mathbf D}
\def\bG{\mathbf G}

\def\bI{\mathbf I}

\def\bW{{\mathbf W}}
\def\bX{{\mathbf X}}

\def\bZ{{\mathbf Z}}

\def\bmu{{\boldsymbol \mu}}

\def\argmin{\mathop{\rm arg\,min}\limits}

\newcommand{\trace}{{\rm trace}}

\newcommand{\eat}[1]{}

\newcommand{\diag}{{\rm diag}}

\newcommand{\thead}[1]{\multicolumn{1}{c}{\textbf{#1}}}
\newcommand{\lhead}[1]{\multicolumn{1}{l}{\textbf{#1}}}

\begin{document}

\title{A Riemannian Approach to Ground Metric Learning \\ for Optimal Transport
}


\author{\IEEEauthorblockN{Pratik Jawanpuria}
\IEEEauthorblockA{
\textit{Microsoft India}\\
pratik.jawanpuria@microsoft.com  
}
\and
\IEEEauthorblockN{Dai Shi}
\IEEEauthorblockA{
\textit{The University of Sydney}\\
dai.shi@sydney.edu.au
}
\and
\IEEEauthorblockN{Bamdev Mishra}
\IEEEauthorblockA{
\textit{Microsoft India}\\
bamdevm@microsoft.com
}
\and
\IEEEauthorblockN{Junbin Gao}
\IEEEauthorblockA{
\textit{The University of Sydney}\\
junbin.gao@sydney.edu.au
}
}

\maketitle

\begin{abstract}

Optimal transport (OT) theory has attracted much attention in machine learning and signal processing applications. OT defines a notion of distance between probability distributions of source and target data points. A crucial factor that influences OT-based distances is the ground metric of the embedding space in which the source and target data points lie. In this work, we propose to learn a suitable latent ground metric parameterized by a symmetric positive definite matrix. We use the rich Riemannian geometry of symmetric positive definite matrices to jointly learn the OT distance along with the ground metric. Empirical results illustrate the efficacy of the learned metric in OT-based domain adaptation.

\end{abstract}


\section{Introduction}

Optimal Transport (OT) \cite{villanioldnew,CompOT} is a mathematical framework for comparing probability distributions by finding the most cost-effective way to transform one distribution into another. It measures the ``distance'' between distributions based on the cost of transporting mass from one point to another \cite{cuturi13}. In machine learning, OT has been applied in areas such as supervised classification \cite{Wloss15}, domain adaptation \cite{courty2016optimal,GaninLempitsky2015}, generative modeling (e.g., Wasserstein GANs) \cite{ArjovskyChintalaBottou2017}, and distribution alignment \cite{alvarez18wordEmb,NIPS2019_9501,tam-etal-2019-optimal,pmlr-v108-janati20a}, offering a principled way to compare and align distributions with minimal assumptions about their structure. The Wasserstein distance, derived from OT, provides a more meaningful metric in high-dimensional settings compared to traditional methods like the Kullback-Leibler divergence. OT is also used for tasks like image registration \cite{FeydyCharlierVialardPeyre2017}, data clustering \cite{Cuturi14b,DesseinPapadakisRouas2018a}, model interpolation \cite{solomon15graphics,bioapp19}, and transfer learning \cite{WangYu2022}.

OT relies heavily on the ground cost metric \cite{Cuturi14a,Kerdoncuff21}, which defines the ``cost'' of transporting mass from one point in a source distribution to another in a target distribution. This cost metric essentially captures how ``far'' points are from each other, and it plays a critical role in how the OT problem is solved, as the goal of OT is to minimize the overall transportation cost based on this metric.  In many applications, the optimal ground cost metric requires domain knowledge to properly capture the relationships between points in the data. Designing this ground cost often requires deep domain expertise, which is not always available. This manual crafting can be time-consuming, and if the metric is poorly designed, it can lead to sub-optimal transport plans and poor performance in downstream tasks. Learning the ground cost from data is an alternative approach that can overcome the limitations of handcrafted metrics, making OT more flexible and applicable across diverse domains without requiring extensive prior knowledge.

This paper motivates ground metric learning in OT. Notably, we jointly learn a suitable underlying ground metric of the embedding space and the transport plan between the given source and target domains. By doing so, the proposed methodology adapts the ground OT cost to better reflect the relationships in the data, which may significantly improve the OT performance. 
Our main contributions are as follows:
\begin{itemize}
    \item We propose a novel ground metric learning based OT formulation in which the latent ground metric is  parameterized by a symmetric positive definite (SPD) matrix $\bA$. Using the rich Riemannian geometry of SPD matrices, we appropriately regularize $\bA$ to avoid trivial solutions. 
    \item We show that the joint optimization over the transport plan $\gamma$ and the SPD matrix $\bA$ can be neatly decoupled in an alternate minimization setting. For a given metric $\bA$, the  transport plan $\gamma$ is efficiently computed via the Sinkhorn method~\cite{knight2008sinkhorn,cuturi13}. Conversely, for a given $\gamma$, optimization over $\bA$ has a closed-form solution. Interestingly, this may be viewed as computing the geometric mean between a pair of SPD matrices under the affine-invariant Riemannian metric \cite{bhatia09a,boumal2023introduction}. 
    \item We evaluate the proposed approach in domain adaptation settings where the source and target datasets have different class  and feature distributions. Our approach outperforms the baselines in terms of generalization performance as well as robustness. 
\end{itemize}





\section{Background and related works}

Let $\X\coloneqq\{\bx_i\}_{i=1}^m$ and $\Z\coloneqq\{\bz_j\}_{j=1}^n$ be independently and identically distributed (i.i.d.) samples of dimension $d$ from distributions $p$ and $q$, respectively. Let $p = \sum_{i=1}^m \bp_i\delta_{\bx_i}$ and $q = \sum_{i=1}^n \bq_j\delta_{\by_i}$ be the empirical distributions corresponding to $p$ and $q$, respectively. Here, $\delta$ denotes the Dirac delta function. We note that $\bp\in\Delta_{m}$ and $\bq\in\Delta_{n}$, where $\Delta_{m} = \{ \bp \in \mathbb{R}_+^{m}: \bp^\top \bone_m = 1\}$. 

The optimal transport (OT) problem \cite{KatoroOT,CompOT} seeks to determine a joint distribution $\gamma$ between the source set $\X$ and the target set $\Z$, ensuring that the marginals of $\gamma$ match the given marginal distributions $p$ and $q$, while minimizing the expected transport cost. The classical OT problem may be stated as 
\begin{equation}\label{eqn:classicalOT}
     \min_{\gamma \in \Gamma(\bp,\bq)}\
 \sum_{i=1}^m\sum_{j=1}^n \gamma_{ij} \bG_{ij},
\end{equation}
where $\Gamma(\bp,\bq)=\{\gamma: \gamma\geq \bzero,\gamma\bone_n = \bp,\gamma^\top\bone_m = \bq\}$. The cost matrix $\bG\in\RR_{+}^{m\times n}$ represents a given ground metric and is computed as $\bG_{ij}=g(\bx_i,\bz_j)$, where $g:\RR^{d}\times \RR^{d}\rightarrow \RR_{+}:(\bx,\bz)\rightarrow g(\bx,\bz)$. Here, the function $g$ formalizes the cost of transporting a unit mass from the source to the target domain. 

The regularized OT formulation with squared Euclidean ground metric may be written as 
\begin{equation}\label{eq:OT}
 \min_{\gamma \in \Gamma(\bp,\bq)}\
 \sum_{i=1}^m\sum_{j=1}^n \gamma_{ij} \|\bx_i-\bz_j\|^2 + \lambda\Omega(\gamma),
\end{equation}
where $\Omega$ is a regularizer on the transport matrix $\gamma$ and $\lambda>0$ is the regularization hyperparameter. In his seminal work, Cuturi \cite{cuturi13} proposed the negative entropy regularizer ($\Omega(\gamma)\coloneqq\sum_{i=1}^m\sum_{j=1}^n \gamma_{ij}\ln(\gamma_{ij})$) and studied its attractive computational and generalization benefits. In particular, (\ref{eq:OT}) with the negative entropy regularizer may be very efficiently solved using the Sinkhorn algorithm \cite{cuturi13,knight2008sinkhorn}. 


It should be noted that (\ref{eq:OT}) employs the (squared) Euclidean distance between the source and target data points. While the squared Euclidean distance may be suitable for spherical data clouds (isotropic distributed), one may employ the (squared) Mahalanobis distance to cater to more general settings, i.e.,
\begin{equation}\label{eq:mahalanobis_OT}
    \min_{\gamma \in \Gamma(\mathbf{p},\mathbf{q})}\ 
    \sum_{i=1}^m\sum_{j=1}^n \gamma_{ij} \|\bx_i-\bz_j\|_{\bA}^2
     +\lambda\Omega(\gamma), 
\end{equation}
where $\bA$ is a given symmetric positive definite (SPD) matrix of size $d\times d$ and $\|\bz\|^2_{\bA} = \bz^\top\bA\bz$. Conceptually, $\bA$ allows to capture the ground metric of the embedding space of the data points. The setting $\bA=\bI$ in Problem (\ref{eq:mahalanobis_OT}) recovers Problem (\ref{eq:OT}). However, obtaining a good (nontrival) $\bA$ for a given problem instance requires domain expertise and it may not be easily available. It the following, we propose a data dependent approach to learn $\bA$ (along with $\gamma$) in an unsupervised setting. 

\section{Proposed approach}
For a given source $\X$ and target $\Z$ datasets, we propose the following formulation to jointly learn the transport plan $\gamma$ and the ground metric $\bA$: 
\begin{equation}\label{eq:reg_mahalanobis_OT}
    \min_{\gamma \in \Gamma(\mathbf{p},\mathbf{q})}\ \min_{
    \bA \succ \bzero
    } \ 
    \sum_{i=1}^m\sum_{j=1}^n \gamma_{ij} \|\bx_i-\bz_j\|_{\bA}^2
     + \Phi(\bA) +\lambda\Omega(\gamma), 
\end{equation}
where the term $\Phi(\bA)$ regularizes the SPD matrix $\bA$. 
We note that Problem (\ref{eq:reg_mahalanobis_OT}) without $\Phi(\bA)$ or with commonly employed regularizers such as $\Phi(\bA) = \|\bA\|_F^2$ or $\Phi(\bA) = {\rm trace}(\bA\bD)$ where $\bD$ is a given (fixed) SPD matrix is not a suitable problem as they lead to a trivial solution with $\bA = \bzero$. 
In this work, we propose $\Phi(\bA) = \langle \bA^{-1},\bD \rangle = \trace{(\bA^{-1}\bD)}$, where $\bD\succ \bzero$ is given.  
Some useful modeling choices of $\bD$ include: $\bD = \bI$ or $\bD = \bX \bX ^\top + \bZ \bZ ^\top$ or $\bD = (\bX \bX ^\top + \bZ \bZ ^\top )^{-1}$, where $\bX = [\bx_1, \bx_2, \ldots, \bx_m]$ and $\bZ = [\bz_1, \bz_2, \ldots, \bz_n]$.  
Below, we provide two motivations for why the term of $\langle \bA^{-1},\bD \rangle$ is interesting. 
\begin{enumerate}
    \item Minimizing the term $\trace{(\bA^{-1}\bD)}$ for $\bA$ only ensures the $\bA^{-1}$ tends to $\bzero$. In contrast, minimizing the term $\sum_{i=1}^m\sum_{j=1}^n \gamma_{ij} \|\bx_i-\bmu_j\|_{\bA}^2$ implies that $\bA$ tends to $\bzero$. Minimizing the sum of both the expressions bounds the solution $\bA$ away from $\bzero$ while keeping the norm of $\bA$ also bounded. 

    \item For a given (fixed) $\gamma$, the optimality conditions of (\ref{eq:reg_mahalanobis_OT}) w.r.t. $\bA$ (discussed in Section~\ref{sec:optimization_method}) provides the following necessary and sufficient condition for optimal $\bA$:
    \begin{equation*}
    \begin{array}{c}
        \bA(\sum_{i=1}^m\sum_{j=1}^n \gamma_{ij}(\bx_i-\bz_j)(\bx_i-\bz_j)^\top) \bA = \bD.
    \end{array}
    \end{equation*}
    We note that $\bA$ which satisfies the above conditions (we discuss this in Section \ref{sec:optimization_method}) ensures that the covariance of the features of the data points will align with that of $\bD$. Hence, setting $\bD = \bI$ implies that $\bA$ promotes the transformed features to become more uncorrelated. 
\end{enumerate}

\section{Optimization algorithm} \label{sec:optimization_method}

It should be noted that Problem (\ref{eq:reg_mahalanobis_OT}) is a minimization problem over the parameters $\gamma$ and $\bA$. We propose to solve it with a minimization strategy that  alternates between the metric learning problem for learing $ \bA$ (for a given $\gamma$) and the OT problem for learning $\gamma$ (for a given $\bA$). This is shown in Algorithm \ref{alg:proposed}. Given $\gamma$, the update for $\bA$ follows from the discussion below and has a closed-form expression. Given $\bA$, the update for $\gamma$ is obtained by solving a OT problem which can be solved by the  {Sinkhorn} algorithm \cite{knight2008sinkhorn,cuturi13}.

\subsection{Metric learning problem: fixing $\gamma$, solve (\ref{eq:reg_mahalanobis_OT}) for $\bA$}  \label{sec:fix_gamma}

In this case, we are interesting in solving the subproblem:
\begin{equation}\label{eq:fix_gamma}
\begin{array}{rl}
   & \min\limits_{
    \bA \succ \bzero
    } \ 
    \sum_{i=1}^m\sum_{j=1}^n \gamma_{ij} \|\bx_i-\bz_j\|_{\bA}^2
     +\langle \bA^{-1},\bD \rangle. \\
    \end{array} 
\end{equation}
Below, we characterize the unique solution of Problem (\ref{eq:fix_gamma}).
\begin{proposition}
    Given $\gamma$, the global optimal $\bA^*$ for Problem (\ref{eq:fix_gamma}) is:
    \begin{equation}
    \begin{array}{ll}
          \bA^* = \bC_{\gamma}^{-1/2}  (\bC_{\gamma}^{1/2} \bD \bC_{\gamma}^{1/2})^{1/2} \bC_{\gamma}^{-1/2},
    \end{array}
    \end{equation}
    where $\bC_\gamma =   \sum_i\sum_j \gamma_{ij}(\bx_i-\bz_j)(\bx_i-\bz_j)^\top$. 
\end{proposition}

    \noindent\textit{Proof.} We first observe that $
    \sum_{i=1}^m\sum_{j=1}^n \gamma_{ij} \|\bx_i-\bz_j\|_{\bA}^2 $ can be written as $    \langle \bA, \sum_i\sum_j \gamma_{ij}(\bx_i-\bz_j)(\bx_i-\bz_j)^\top \rangle$. Consequently, the objective function is rewritten as $ \langle \bA, \bC_\gamma\rangle +  \langle \bA^{-1}, \bD \rangle$, where $\bC_\gamma =   \sum_i\sum_j \gamma_{ij}(\bx_i-\bz_j)(\bx_i-\bz_j)^\top$. The objective function is convex in $\bA$. Furthermore, the characterization of the first-order KKT conditions for (\ref{eq:fix_gamma}) leads to the condition $\bC_{\gamma} = \bA^{-1} \bD \bA^{-1}$ which needs to be solved for a SPD $\bA$. This is equivalent to the condition $\bA \bC_{\gamma} \bA = \bD$. From \cite[Exercise~1.2.13]{bhatia09a}, this quadratic equation is called the Riccati equation and employs a unique solution for SPD matrices $\bC_{\gamma}$ and $\bD$. The solution is obtained by multiplying $\bC_{\gamma}^{-1/2}$ to both the left-hand and right-hand sides and taking the principal square root. This completes the proof.\hfill $\square$


A novel viewpoint of solving (\ref{eq:fix_gamma}) is further explored in \cite{zadeh16} that exploits the affine-invariant Riemannian geometry of SPD matrices \cite{bhatia09a,absil2008optimization}. From the viewpoint of Riemannian geometry, the objective function $\langle \bA, \bC_{\gamma} \rangle + \langle \bA^{-1}, \bD\rangle$ on the space of the Riemannian manifold of SPD matrices is viewed as computing the geometric mean between two SPD matrices: $\bC_{\gamma}^{-1}$ and $\bD$. The geometric mean corresponds to the midpoint of the Riemannian geodesic curve connecting $\bC_{\gamma}^{-1}$ and $\bD$ \cite{bhatia09a}. In fact, the function $\langle \bA, \bC_{\gamma} \rangle + \langle \bA^{-1}, \bD\rangle$ is also geodesic convex on the SPD manifold. In scenarios, where $d > m+n$, note that $\bC_\gamma$ is symmetric positive semi-definite. To this end, we add a regularization term $\epsilon\bI$ to $\bC_{\gamma}$, where $\epsilon$ is a small positive scalar.


\begin{algorithm}[t]
\caption{Algorithm for (\ref{eq:reg_mahalanobis_OT}).}\label{alg:proposed}
\begin{algorithmic}
\State \textbf{Input:} Source and target data points $\{\bx_i\}_{i=1}^m$ and $\{\bz_j\}_{j=1}^n$, respectively.
\State \textbf{Initialize:} $\gamma_0 = \bp \bq^\top$ to be a uniform joint probability matrix of size $m\times n$.
\State \textbf{for} $t = 1, \ldots, l$
\State\quad $\bA^{t} = \bC_{\gamma^{t-1}}^{-1/2}  (\bC_{\gamma^{t-1}}^{1/2} \bD \bC_{\gamma^{t-1}}^{1/2})^{1/2} \bC_{\gamma^{t-1}}^{-1/2}$, 
\State \qquad \quad where $\bC_{\gamma^{t-1}} = \sum_i\sum_j \gamma^{t-1}_{ij}(\bx_i-\bz_j)(\bx_i-\bz_j)^\top$.

\State\quad $\gamma^{t} = \texttt{Sinkhorn}(\bC_{\bA^{t}}, \bp, \bq)$,  
\State \qquad \quad where $\bC_{\bA^{t}}$ is computed using (\ref{eqn:CA}).
\State \textbf{end for}
\State \textbf{Output}: Joint probability matrix $\gamma^l$ and the metric $\bA^l$.
\end{algorithmic}
\end{algorithm}

\subsection{OT problem: fixing $\bA$, solve (\ref{eq:reg_mahalanobis_OT}) for $\gamma$} 
In this case, we need to solve the subproblem: 
\begin{equation}\label{eq:fix_mu_A}
\begin{array}{l}
     \min\limits_{\gamma \in \Gamma(\mathbf{p},\mathbf{q})} \ 
    \langle \gamma, \bC_{\bA}\rangle
     + \lambda\Omega(\gamma),
\end{array}
\end{equation}
where $\bC_{\bA}(i,j)=\|\bx_i-\bz_j\|_{\bA}^2$. We compute $\bC_{\bA}$ efficiently as 
\begin{equation}\label{eqn:CA}
\bC_{\bA} = \diag(\bX^\top \bA \bX) \bone_n^\top + \bone_m \diag(\bZ^\top \bA \bZ)^\top - 2 \bX^\top \bA \bZ,
\end{equation}
where $\diag$ extracts the diagonal element of a square matrix as a column vector. Problem (\ref{eq:fix_mu_A}) is viewed as a instance of the Problem (\ref{eq:OT}) but now with the cost matrix $\bC_{\bA}$. As discussed, we solve it by the {Sinkhorn} algorithm \cite{cuturi13}.

\section{Experiments}
We empirically study our approach in domain adaptation scenarios \cite{courty2016optimal,Courty17domAda}, an important application area of optimal transport. 
In our experiments, we focus on evaluating the utility of the proposed joint learning of transport plan $\gamma$ and the ground metric $\bA$ against OT baselines where the ground metric is pre-determined. 

\subsection{Barycentric projection for domain adaptation}\label{subsec:barycentric}
Given a supervised source dataset and an unlabeled target dataset, the aim of domain adaptation is to use source supervision to correctly classify the target instances. If the source and target datasets are from the same domain (with same distribution of features and labels), then no \textit{adaptation} is required and we may use source instances directly. However, if the label (and/or feature) distribution of the source set and the target set differ, then we require adapting the source instances to the target domain. 

Optimal transport (OT) provides a principled approach for comparing the source and the target datasets (and thus their underlying distributions). In particular, the learned transport plan $\gamma$ can be used to \textit{transport} the source points appropriately into the target domain. This can be done efficiently using the barycentric mapping \cite{CompOT}. For both (\ref{eq:mahalanobis_OT}) and the proposed (\ref{eq:reg_mahalanobis_OT}) problems, the barycentric mapping of a source point $\bx_i$ into the target domain is given by 
\begin{equation}\label{eqn:barycenteric_mapping}
    \hat{\bx}_i \coloneqq \argmin_{\bz\in\RR^d} \sum_{j=1}^n \gamma_{ij} \|\bz-\bz_j\|_{\bA}^2 = \sum_{j=1}^n \bz_j\gamma_{ij}/\bp_i. 
\end{equation}
The barycentric mapping (\ref{eqn:barycenteric_mapping}) maps the $i$-th source instance $\bx_i$ to $\hat{\bx}_i$, which is a weighted average of the target set instances. 
The weight $\gamma(i,j)/\bp_i$ denotes the conditional distribution of the target instance $\bz_j$ given the source instance $\bx_i$. 

\textbf{Inference on the target set.} 
Given a labeled source instance $\{\bx_i,y_i\}$, the barycentric projection (\ref{eqn:barycenteric_mapping}) provides a mechanism to obtain its corresponding instance $\{\hat{\bx}_i,y_i\}$ in the target domain. Thus, instead of directly using the source points, their barycentric mappings could be used to classify the target set instances for domain adaptation scenarios. In this work, we employ a 1-Nearest Neighbor (1-NN) classifier for classifying the target instances \cite{courty2016optimal,gurumoorthy21a}. The 1-NN classifier is parameterized by the barycentric mappings of the labeled source instances.

\subsection{Experimental setup}
\textbf{Datasets.} We conduct experiments using the Caltech-Office and MNIST datasets. 
\begin{itemize}
    \item MNIST \cite{lecun1998mnist} is a collection of handwritten digits. It consists of two different image sets of sizes 60,000 and 10,000, respectively. Each image is labeled with a digit from 0 to 9 and has dimension $28\times 28$ pixels. 
    \item Caltech-Office \cite{saenko2010adapting} includes images from four distinct domains: Amazon (online retail), the Caltech image dataset, DSLR (high-resolution camera images), and Webcam (webcam images). These domains differ in various aspects such as background, lighting conditions, and noise levels. The dataset comprises 958 images from Amazon (A), 1123 from Caltech (C), 157 from DSLR (D), and 295 from Webcam (W). Each domain can serve as either the source or target domain. Thus, there are twelve adaptation tasks, one corresponding to every source-target domains pairs (e.g., $A\rightarrow D$ implies A is source and D is target). We utilize DeCAF6 features to represent the images \cite{donahue14a}.
\end{itemize}

\textbf{Source and target sets.} For both MNIST and Caltech-Office, we perform multi-class classification in the target domain using labeled data exclusively from the source domain (as discussed in Section~\ref{subsec:barycentric}). The source and target sets are created as follows for the two datasets:
\begin{itemize}
    \item MNIST: Following \cite{gurumoorthy21a}, the source set $\X$ is created such that every label has uniform distribution. The target training and test sets, $\Z_{t}$ and $\Z_e$, respectively, are created such that they have a skewed distribution for a chosen class $c$. The data points corresponding to class $c$ constitute $w\%$ of the target sets and the other classes uniformly constitute the remaining $(100-w)\%$. We experiment with $z=\{10,20,30,40,50\}$. Setting $z=10$ implies uniform label distributions in the target sets (same as $\X$). However, both $\Z_t$ and $\Z_e$ have different label distribution than $\X$ when $z>10$. 
    The chosen class $c$ is varied from digits 0 to 9 in our experiments for every $z$. 
    For each run, we sample $m=500$ points from the 10K set for $\X$ and sample $n=500$ instances from the 60K set for both $\Z_t$ and $\Z_e$. We ensure that $Z_t \cap Z_e = \{\}$. 
    \item Caltech-Office: For each task, we randomly select ten images per class from the source domain to create the source set $\X$ (for source domain D, we select eight per class due to small sample size). The target domain is divided equally into training ($\Z_t$) and test ($\Z_e$) sets \cite{courty2016optimal}. 
\end{itemize}

\textbf{Training and evaluation.} We use $\{\bx_i\}_{i=1}^m\in\X$ and $\{\bz_j\}_{i=1}^n\in\Z_t$ to learn the transport plan $\gamma$ for all the algorithms and the ground metric $\bA$ for the proposed approach. 
The hyperparameter $\lambda$ for all the algorithms is tuned using the accuracy of the corresponding 1-NN classifier on the target train set $\Z_t$. We report the accuracy obtained on the target test set $\Z_e$ with tuned $\lambda$. All experiments are repeated five times with different random seeds and averaged results are reported.

\textbf{Baselines.}
We compare our proposed approach (Algorithm~\ref{alg:proposed}) with $\bD=\bI$ against the Mahalanobis distance based OT (\ref{eq:mahalanobis_OT}) where the metric is fixed. In particular, we experiment with the three given (fixed) metric baselines: 
\begin{enumerate}
    \item OT$_{\bI}$: it employs $\bA = \bI$ in (\ref{eq:mahalanobis_OT}), i.e., the squared Euclidean cost.
    \item OT$_{\bW^{-1}}$: it employs $\bA = \bW^{-1}$ in (\ref{eq:mahalanobis_OT}) where $\bW=[\bX,\bZ][\bX,\bZ]^\top$. Such a choice of $\bA$ leads to whitening/decorrelation of the data. 
    \item OT$_{\bW}$: as a third baseline, we also explore $\bA = \bW$ in (\ref{eq:mahalanobis_OT}), where $\bW=[\bX,\bZ][\bX,\bZ]^\top$.
\end{enumerate}

\subsection{Results and discussion}
\textbf{MNIST.} Table~\ref{table:mnist} reports the generalization performance obtained by different methods on the target domains of MNIST. We observe that our approach is robust to the skew present in the target domain. In particular, when skew percentage is high (i.e., the target distribution is quite different from the source distribution) our approach outperforms the three baselines methods. As noted earlier, $z=10$ implies the label distribution in the target set is same as the label distribution in source set. Hence, $z=10$ setting does not require any domain adaptation and we observe that regularized OT with squared Euclidean cost (OT$_{\bI}$) performs the best. We also note that OT$_{\bW^{-1}}$ and OT$_{\bW}$ perform poorly, highlighting the difficulties in obtaining a good hand-crafted ground metric $\bA$. 

\textbf{Caltech-Office.} Table~\ref{table:officeCaltech} reports the generalization performance obtained by different methods on the twelve adaptation tasks of the Caltech-Office dataset. We observe that the proposed approach obtains the best overall result, obtaining best performance in several tasks. We also remark that the performance of all the three baselines are similar to each other. While the baselines obtain best performance in multiple tasks, we interestingly note that our approach is a close second (or third in the case of $W\rightarrow D$) in the corresponding tasks. However, in the tasks where the proposed approach obtains the best accuracy, it outperforms the baselines by some margin, underlying   the significance of ground metric learning for OT.


\begin{table}
\caption{Average accuracy on the target domains of MNIST. The label distribution in the source set is uniform and in the target set is skewed. Our approach is robust to the label distribution shifts, outperforming baselines in the challenging settings with higher skew.}
\label{table:mnist}
\centering
\begin{tabular}{ccccc}
\toprule
\thead{Skew (\%)} & \thead{OT$_{\bI}$} & \thead{OT$_{\bW}$} & \thead{OT$_{\bW^{-1}}$} & \thead{Proposed}\\
\midrule
$10$ & $\mathbf{85.24}$ & $66.66$ & $46.72$ & $82.44$\\
$20$ & $\mathbf{83.72}$ & $66.28$ & $46.69$ & $82.28$\\
$30$ & $79.91$ & $66.77$ & $46.91$ & $\mathbf{82.31}$\\
$40$ & $74.57$ & $67.11$ & $46.44$ & $\mathbf{82.39}$\\
$50$ & $73.10$ & $67.56$ & $46.26$ & $\mathbf{81.49}$\\
\bottomrule
\end{tabular}
\end{table}

\begin{table}
\caption{Average accuracy on the target domains of the Caltech-Office dataset. Our approach obtains the best overall results, signifying the importance of ground metric learning for OT.} \label{table:officeCaltech}
\centering
\begin{tabular}{lcccc}
\toprule
\lhead{Task} & \thead{OT$_{\bI}$} & \thead{OT$_{\bW}$} & \thead{OT$_{\bW^{-1}}$} & \thead{Proposed}\\
\midrule
$A\rightarrow C$ & $\mathbf{84.21}$ & $79.93$ & $80.64$ & $83.39$\\
$A\rightarrow D$ & $80.00$ & $\mathbf{81.52}$ & $78.99$  & $80.00$\\
$A\rightarrow W$ & $76.76$ & $76.49$ & $77.16$ & $\mathbf{79.59}$\\
$C\rightarrow A$ & $87.28$ & $86.25$ & $87.15$   & $\mathbf{87.79}$\\
$C\rightarrow D$ & $76.96$ & $76.71$ & $76.71$   & $\mathbf{79.75}$\\
$C\rightarrow W$ & $69.73$ & $\mathbf{73.38}$ & $70.68$   & $72.57$\\
$D\rightarrow A$ & $85.52$ & $\mathbf{87.54}$ & $85.48$  & $87.24$\\
$D\rightarrow C$ & $80.89$ & $79.86$ & $80.75$  & $\mathbf{83.14}$\\
$D\rightarrow W$ & $\mathbf{95.14}$ & $92.70$ & $94.32$  & $95.00$\\
$W\rightarrow A$ & $79.74$ & $\mathbf{81.50}$ & $78.12$  & $81.41$\\
$W\rightarrow C$ & $75.76$ & $73.65$ & $74.19$  & $\mathbf{78.47}$\\
$W\rightarrow D$ & $91.65$ & $93.42$ & $\mathbf{93.67}$  & $93.16$\\
\midrule
Average & $81.97$ & $81.91$ & $81.49$  & $\mathbf{83.46}$\\
\bottomrule
\end{tabular}
\end{table}

\section{Conclusion}
In this work, we proposed a novel framework for ground metric learning in optimal transport (OT) by leveraging the Riemannian geometry of symmetric positive definite (SPD) matrices. By jointly learning the transport plan and the ground metric, our approach adapts the ground cost metric to better reflect the relationships in the data. Thus, our approach enhances the flexibility and applicability of OT, making it suitable for tasks without extensive domain knowledge. 
Our algorithm efficiently optimizes two convex problems alternatively: a metric learning problem and an OT problem. The metric learning problem, in particular, is solved in closed-form and is related to computing the geometric mean of a pair of SPD matrices under the Riemannian metric. 
Empirically, our method consistently outperforms OT baselines in domain adaptation benchmarks, underscoring the significance of learning a suitable ground metric for OT applications.

\bibliographystyle{IEEEtran}
\bibliography{references}
\end{document}